%% file: main.tex
\documentclass[11pt]{article}
\usepackage{fullpage}
\usepackage{amsmath,amsfonts,amsthm,amssymb}
\usepackage{url}

\usepackage{color}
\usepackage[usenames,dvipsnames,svgnames,table]{xcolor}
\usepackage[colorlinks=true, linkcolor=red, urlcolor=blue, citecolor=blue]{hyperref}

\usepackage{booktabs}
\usepackage{algorithm}
\usepackage{algorithmic}
\usepackage{comment}
\usepackage{mathtools}
\usepackage{microtype}
\usepackage{caption}
\usepackage{float}
\usepackage{graphicx}
\usepackage{multirow}
\usepackage{mathrsfs}
\usepackage{newfloat}
\usepackage{relsize}
\usepackage{dsfont}
\usepackage{booktabs,siunitx,caption}

\usepackage[utf8]{inputenc} 
\usepackage[T1]{fontenc}    
\usepackage{url}            
\usepackage{booktabs}       
\usepackage{nicefrac}       

\usepackage{thm-restate}
\usepackage{bussproofs}
\usepackage{mdframed}
\usepackage{mdframed}
\usepackage{listings}
\usepackage{tikz}
\usepackage{fancybox}
\usepackage{verbatim}
\usepackage{dsfont}
\usepackage{subcaption}
\usepackage{multirow}
\usepackage{multicol}

\usepackage{algorithm}
\usepackage{algorithmic}
 
\usepackage{numprint}
\usepackage{scalerel}

\usepackage{listings}
\usepackage{paralist}
\usepackage{multirow}
\usepackage{soul}

\usetikzlibrary{tikzmark}
\definecolor{codegreen}{rgb}{0,0.6,0}
\definecolor{codegray}{rgb}{0.5,0.5,0.5}
\definecolor{codepurple}{rgb}{0.58,0,0.82}
\definecolor{backcolour}{rgb}{0.95,0.95,0.92}
 
\lstdefinestyle{codestyle}{
    backgroundcolor=\color{backcolour},   
    commentstyle=\color{codegreen},
    keywordstyle=\color{blue},
    numberstyle=\tiny\color{codegray},
    stringstyle=\color{codepurple},
    basicstyle=\ttfamily\footnotesize,
    breakatwhitespace=false,         
    breaklines=true,                 
    captionpos=b,                    
    keepspaces=true,                 
    numbers=left,                    
    numbersep=5pt,                  
    showspaces=false,                
    showstringspaces=false,
    showtabs=false,                  
    tabsize=1
}
\input{commands.tex}

\let\citep\cite

\title{Intra-Processing Methods for Debiasing Neural Networks}

\author{
Yash Savani\\
Abacus.AI\\
\texttt{yash@abacus.ai}\\
\and
Colin White\\
Abacus.AI\\
\texttt{colin@abacus.ai}\\
\and 
Naveen Sundar Govindarajulu\\
RAIR Lab RPI\\
\texttt{naveensundarg@gmail.com}\\
}

\date{}

\begin{document}

\maketitle

\begin{abstract}

\input{abstract.tex}

\end{abstract}

\section{Introduction} \label{sec:intro}
\input{intro.tex}

\section{Related Work}\label{sec:related_work}

\input{related.tex}

\section{Broader Impact} \label{sec:impact}

\input{impact}

\section{Preliminaries}\label{sec:prelim}

\input{prelim.tex}

\section{Methodology}\label{sec:method}

\input{method}

\section{Experiments}\label{sec:experiments}

\input{experiments.tex}

\section{Conclusion}\label{sec:conclusion}
\input{conclusion.tex}

\clearpage

\bibliography{main}
\bibliographystyle{plain}

\clearpage
\appendix

\input{appendix_experiments}

\end{document}

%% file: commands.tex
\newcommand{\arity}[1]{\mathitf{arity}[f]}

\usepackage{amsmath}

\newcommand{\tocite}[1]{\hl{cite}}

\newcommand{\mar}{\delta}

\newcommand{\R}{\mathbb{R}}

\newcommand{\ind}{\mathbb{I}}

\newcommand{\Y}{\mathcal{Y}}

%% file: abstract.tex
As deep learning models become tasked with more and more decisions
that impact human lives, such as criminal recidivism, loan repayment,
and face recognition for law enforcement,
bias is becoming a growing concern. 
Debiasing algorithms are typically split into three paradigms:
pre-processing, in-processing, and post-processing.
However, in computer vision or natural language applications, 
it is common to start with a large generic model and then 
fine-tune to a specific use-case.
Pre- or in-processing methods would require retraining the entire model from scratch,
while post-processing methods only have black-box access to the model, so they
do not leverage the weights of the trained model.
Creating debiasing algorithms specifically for this fine-tuning use-case has largely been neglected.

In this work, we initiate the study of a new paradigm in debiasing research,
\emph{intra-processing}, which sits between in-processing and post-processing methods.
Intra-processing methods are designed specifically to debias large models which
have been trained on a generic dataset and fine-tuned on a more specific task.
We show how to repurpose existing in-processing methods for this use-case,
and we also propose three baseline algorithms: random perturbation, 
layerwise optimization, and adversarial fine-tuning.
All of our techniques can be used for all popular group fairness measures such as equalized odds or statistical parity difference.
We evaluate these methods across three popular datasets from the AIF360 toolkit,
as well as on the CelebA faces dataset.
Our code is available at \url{https://github.com/abacusai/intraprocessing\_debiasing}.

%% file: intro.tex
The last decade has seen a huge increase in applications 
of machine learning in a wide variety of domains such as
credit scoring, fraud detection, hiring decisions,
criminal recidivism, loan repayment, face recognition, 
and so on~\cite{mukerjee2002multi, bogen2018help, ngai2011application, barocas2017fairness, learned2020facial}.
The outcome of these algorithms are impacting the
lives of people more than ever.
There are clear advantages in the automation of classification
tasks, as machines can quickly process thousands of datapoints with
many features. However, algorithms are susceptible to bias towards
individuals or groups of people from a variety of 
sources~\cite{o2016weapons, executive2014big, executive2016big, wilson2019predictive,buolamwini2018gender,joo2020gender,wang2020towards}.

For example, facial recognition algorithms are currently being used by the 
US government to match application photos from people applying for visas and 
immigration benefits, to match mugshots, and to match photos as people cross the border 
into the USA~\citep{hao2019us}. 
However, recent studies showed that many of these algorithms exhibit bias 
based on race and gender~\citep{grother2019face}. 
For example, some of the algorithms were 10 or 100 times more 
likely to have false positives for Asian or Black people, compared to white people. 
When used for law enforcement, it means that a Black or Asian person is more likely 
to be arrested and detained for a crime they didn’t commit~\citep{allyn2020computer}. 

Motivated by the discovery of biased models in real-life applications,
the last few years has seen a huge
growth in the area of fairness in machine learning.
Dozens of formal definitions of fairness have been proposed~\cite{narayanan2018translation},
and many algorithmic techniques have been developed for 
debiasing according to these definitions~\cite{verma2018fairness}.
Many debiasing algorithms fit into one of three categories:
pre-processing, in-processing, or post-processing~\citep{d2017conscientious, bellamy2018ai}.
Pre-processing techniques make changes to the data itself, in-processing techniques
are methods for training machine learning models tailored to making fairer models,
and post-processing techniques modify the final predictions outputted by a 
(biased) model.

However, as datasets become larger and training becomes more computationally intensive,
especially in the case of computer vision and natural language processing,
it is becoming increasingly more common in applications to start with a very large
pretrained model, and then fine-tune for the specific 
use-case~\citep{ouyang2016factors, kading2016fine, chi2017thyroid, too2019comparative}.
In fact, PyTorch offers several pretrained models, all of which have been
trained for dozens of GPU hours on ImageNet~\citep{pytorch}.
Pre-, in-, and post-processing debiasing methods are of little
help here: pre- and in-processing methods would require retraining the 
entire model from scratch, 
and post-processing methods would not make use of the full power of the model.

\begin{figure*}
\centering %
\includegraphics[width=0.98\textwidth]{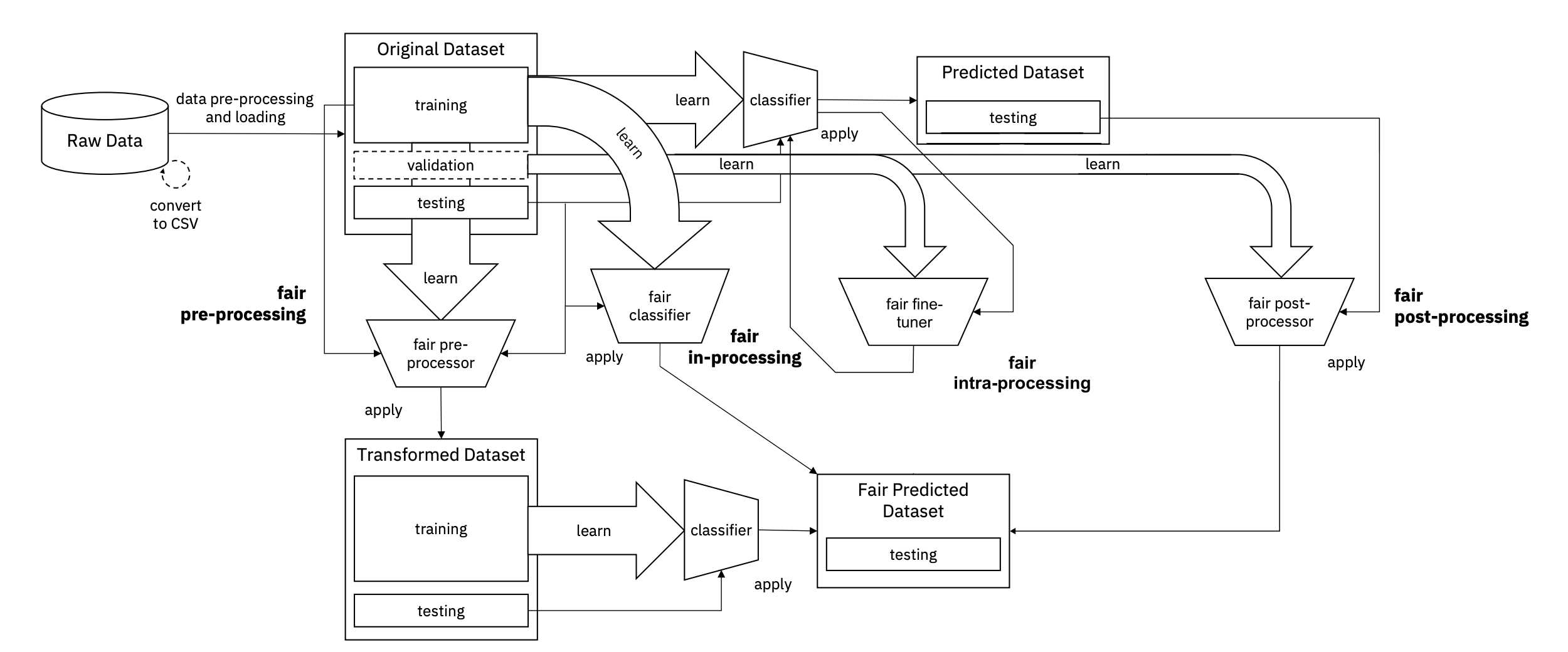}
\caption{
An overview of the debiasing algorithm classes 
(inspiration from the figure in~\cite{bellamy2018ai}). 
The classical classes are the pre-processing, the in-processing and the post-processing algorithms. We introduce the novel intra-processing algorithm class.
}
\label{fig:fairai_overview}
\end{figure*}

In this work, we initiate the study of \emph{intra-processing} methods
for debiasing neural networks. An intra-processing method is defined as an
algorithm which has access to a trained model and a dataset (which typically
differs from the original training dataset), and outputs a new model which
gives debiased predictions on the target task
(typically by updating or augmenting the weights of the original model). To see an overview of all the fairness debiasing algorithm classes, see Figure \ref{fig:fairai_overview}.
We propose three different intra-processing baseline algorithms, and we also show how
to repurpose a popular in-processing algorithm~\citep{zhang2018mitigating} 
to the intra-processing setting.
All of the algorithms we study work for any group fairness measure and any objective
which trades off accuracy with bias.

Our first baseline is a simple random perturbation algorithm, 
which iteratively adds multiplicative noise to the weights of the neural network 
and then picks the perturbation which maximizes the chosen objective.
Our next baseline optimizes the weights of each layer using GBRT~\citep{friedman2001greedy}.
Finally, we propose adversarial methods for fine-tuning.
Adversarial training was recently used as an in-processing method for
debiasing~\citep{zhang2018mitigating}, by training a critic model to predict
the protected attribute of datapoints, to ensure that the predictions are not
correlated with the protected attribute. 
We modify this approach to fit the intra-processing setting,
and we also propose a new, more direct approach which trains a critic to 
\emph{directly measure the bias} of the model weights,
which gives us a differentiable proxy for bias, enabling the use of
gradient descent for debiasing.

We compare the four above techniques with three post-processing 
algorithms from prior work: 
reject option classification~\cite{kamiran2012decision},
equalized odds post-processing~\citep{hardt2016equality},
and calibrated equalized odds post-processing~\citep{pleiss2017fairness}.
We run experiments with three fairness datasets from AIF360~\citep{bellamy2018ai},
as well as the CelebA dataset~\citep{celeba}, with three popular fairness definitions.
We show that intra-processing is much more effective than post-processing for the fine-tuning
use case.
We also show that the difficulty of intra-processing and post-processing debiasing is highly 
dependent on the initial conditions of the original model.
In particular, given a neural network trained to optimize accuracy, the
variance in the amount of bias of the trained model is much higher than the
variance in the accuracy, with respect to the random seed used for initializing 
the weights of the original model.
%
Fairness research (and machine learning research as a whole) has seen a huge increase
in popularity, and recent papers have highlighted the need for fair and reproducible
results~\cite{schelter2019fairprep, bellamy2018ai}.
To facilitate best practices, we run our experiments on the AIF360 toolkit~\cite{bellamy2018ai}
and open source all of our code.

\paragraph{Our contributions.} We summarize our main contributions below.
\begin{itemize} 
    \item We initiate the study of intra-processing algorithms for 
    debiasing ML models. This framework sits in between in-processing and
    post-processing algorithms, 
    and is realistic for many fine-tuning use cases.
    \item We study the nature of intra-processing techniques for debiasing neural networks, showing that the problem is sensitive to the initial conditions of the original model.
    \item We propose three intra-processing algorithms, 
    and we show how to repurpose popular in-processing algorithms into the intra-processing setting. 
    We compare all algorithms across a variety of group fairness constraints and
    datasets against three post-processing algorithms.
\end{itemize}

%% file: related.tex
\paragraph{Debiasing overview.}
There is a surging body of research on bias and fairness in machine learning. 
There are dozens of types of bias that can arise~\cite{mehrabi2019survey},
and many formal definitions of fairness have been proposed~\cite{narayanan2018translation}.
Popular definitions include statistical 
parity/demographic parity~\cite{dwork2012fairness, kusner2017counterfactual}, 
equal opportunity (a subset of equalized odds)~\cite{NIPS2016_6374},
and average absolute odds~\cite{bellamy2018ai}.
For an overview of fairness definitions and techniques, see~\citep{bellamy2018ai,verma2018fairness}.
Recently, the AIF360 toolkit was established to facilitate best practices in debiasing 
experiments~\citep{bellamy2018ai}.
A meta-algorithm was also recently developed for in-processing debiasing by reducing
fairness measures to convex optimization problems~\citep{celis2019classification}.
Another work treats debiasing as an empirical risk minimization problem~\cite{donini2018empirical}.
Yet another work adds the fairness constraints as regularizers in the machine learning
models~\citep{berk2017convex}.
A recent work adaptively samples the training dataset during optimization in order to reduce
disparate performance~\citep{abernethy2020adaptive}.
There is also prior work using adversarial learning to debias 
algorithms~\citep{zhang2018mitigating}.
To the best of our knowledge, no prior work has designed an intra-processing algorithm using
adversarial learning.

\paragraph{Post-processing methods.}
Many post-processing debiasing algorithms have been proposed, which all assume black-box
access to a biased classifier~\citep{kim2019multiaccuracy, celeba, NIPS2016_6374, pleiss2017fairness, kamiran2012decision}.
We describe a few of these techniques in Section~\ref{sec:experiments}.
Currently, most of these techniques have only been established for specific 
fairness measures.
In natural language processing, there is recent work which decreases the bias 
present in pretrained models such as BERT or ELMo~\citep{liang2020towards},
however, these techniques were not shown to work on tabular or computer vision datasets.

\paragraph{Debiasing in computer vision.}
A recent work studied methods for debiasing CelebA with respect to 
gender~\citep{wang2020towards}. However, the study was limited to in-processing techniques
focusing on achieving the highest accuracy in light of distribution shift with respect
to correlations present in the data.
Another work considered counterfactual sensitivity analysis for detecting bias on the CelebA
dataset, using the latent space in GANs~\cite{denton2019detecting}.
There is also recent work in measuring and mitigating bias in facial recognition software and
related computer vision 
tasks~\citep{buolamwini2018gender, wilson2019predictive, raji2020saving, learned2020facial}.


%% file: impact.tex
Deep learning algorithms are more prevalent now than ever before.
The technology is becoming integrated into our society,
and is being used in high-stakes applications such as
criminal recidivism, loan repayment, and hiring 
decisions~\cite{mukerjee2002multi, bogen2018help, ngai2011application, barocas2017fairness}.
It is also becoming increasingly evident that many of these algorithms
are biased from various 
sources~\cite{o2016weapons, executive2014big, executive2016big}.
Using technologies that makes prejudiced decisions for life-changing events 
will only deepen the divides that already exist in society.
The need to address these
issues is higher than ever~\cite{bellamy2020message, oliver2020protesting}.

Our work seeks to decrease the negative effects that biased deep learning algorithms
have on society. Intra-processing methods, which work for any group fairness
measure, will be applicable to large existing deep learning models,
since the networks need not be retrained from scratch.
We present simple techniques (random perturbation) as well
as more complex and strong techniques (adversarial fine-tuning) that work with most existing deep learning architectures.
Our study on the nature of our proposed intra-processing debiasing methods compared to prior work,
may also facilitate future work on related debiasing algorithms.

Algorithms to mitigate bias such as the ones in our paper are an important part of creating
inclusive and unbiased technology. However, as pointed out in other work~\cite{denton2019detecting},
bias mitigation techniques must be part of a larger, 
socially contextualized project to examine the
ethical considerations in deploying facial analysis models.

%% file: prelim.tex
In this section, we give notation and definitions used throughout the paper. 
Given a dataset split into three parts,
$\mathcal{D}_{\text{train}},~
\mathcal{D}_{\text{valid}},~\mathcal{D}_{\text{test}}$,
let $(\pmb{x}_i, Y_i)$ denote one datapoint, where
$\pmb{x}_i\in \mathbb{R}^d$ contains $d$ features including one binary protected
attribute $A$ (e.g., identifying as female or not identifying as female),
and $Y_i\in \{0,1\}$ is the label.
Denote the value of the protected feature for $\pmb{x}_i$ as $a_i$.
We denote a trained neural network by a function 
$f_\theta:\R^d\rightarrow [0,1]$, where $\theta$ denotes the trained
weights.
We often denote $f_\theta(\pmb{x}_i)=\hat Y_i$, the output predicted probability for datapoint
$\pmb{x}_i.$
Finally, we refer to the list of labels in a dataset $\mathcal{D}$ as $\Y.$

\paragraph{Fairness measures.}
We now give an overview of group fairness measures used in
this work. 
Given a dataset $\mathcal{D}$ with labels $\Y$, 
protected attribute $A$, and a list of predictions
$\hat{\Y}=[f_\theta(\pmb{x}_i)\mid (\pmb{x}_i, Y_i)\in\mathcal{D}]$ from some neural network
$f_\theta$, we define the true positive and false positive rates as
\begin{align*}
\mathit{TPR}_{A=a}(\mathcal{D}, \hat{\Y})
=\frac{\left|\{i\mid \hat Y_i=Y_i=1,a_i=a\}\right|}{\left|\{i\mid \hat Y_i=Y_i=1\}\right|}
= P_{(\pmb{x}_i,Y_i)\in \mathcal{D}}(\hat Y_i=1\mid a_i=a, Y_i=1),\\
\mathit{FPR}_{A=a}(\mathcal{D}, \hat{\Y})
=\frac{\left|\{i\mid \hat Y_i=1,Y_i=0,a_i=a\}\right|}{\left|\{i\mid \hat Y_i=1,Y_i=0\}\right|}
= P_{(\pmb{x}_i,Y_i)\in \mathcal{D}}(\hat Y_i=1\mid a_i=a, Y_i=0),\\
\end{align*}

\emph{Statistical Parity Difference (SPD)}, or demographic
parity difference~\cite{dwork2012fairness, kusner2017counterfactual}, 
measures the difference in the probability of a positive outcome
between the protected and unprotected groups. Formally,

\begin{equation*}
\mathit{SPD}(\mathcal{D}, \hat{\Y},A) = 
P_{(\pmb{x}_i,Y_i)\in \mathcal{D}}(\hat Y_i=1\mid a_i=0)
-P_{(\pmb{x}_i,Y_i)\in \mathcal{D}}(\hat Y_i=1\mid a_i=1).
\end{equation*}

\emph{Equal opportunity difference (EOD)}~\cite{NIPS2016_6374} 
measures the difference in TPR for the protected
and unprotected groups.
Equal opportunity is identical to \emph{equalized odds} 
in the case where the protected feature and labels are binary.
Formally, we have

\begin{equation*}
\mathit{EOD}(\mathcal{D}, \hat{\Y},A) = \mathit{TPR}_{A=0}(\mathcal{D}, \hat{\Y}) 
- \mathit{TPR}_{A=1}(\mathcal{D}, \hat{\Y}).
\end{equation*}

\emph{Average Odds Difference (AOD)}~\cite{bellamy2018ai} is
defined as the average of the difference in the false positive
rates and true positive rates for unprivileged and
privileged groups. Formally,
\begin{equation*}
\mathit{AOD}(\mathcal{D}, \hat{\Y},A) =
\frac{(\mathit{FPR}_{A=0}(\mathcal{D}, \hat{\Y})   -  \mathit{FPR}_{A=1}(\mathcal{D}, \hat{\Y}) ) +   (\mathit{TPR}_{A=0}(\mathcal{D}, \hat{\Y})   -  \mathit{TPR}_{A=1}(\mathcal{D}, \hat{\Y}) )}{2}.
\end{equation*}

\paragraph{Optimization techniques.}
Zeroth order (non-differentiable) optimization is used when the objective
function is not differentiable (as is the case for most definitions of group fairness).
This is also called black-box optimization.
Given an input space $W$ and an objective function $\mu$, zeroth order optimization
seeks to compute $w^\ast=\text{arg}\min_{w\in W} \mu (w)$.
Leading methods for zeroth order optimization when function queries are expensive
(such as optimizing a deep network) include
gradient-boosted regression trees (GBRT)~\cite{friedman2001greedy, mason2000boosting}
and Bayesian optimization (BO)~\cite{gpml, frazier2018tutorial, snoek2012practical},
however BO struggles with high-dimensional data.
In contrast, first-order optimization is used when it is possible to take the derivative
of the objective function. Gradient descent is an example of a first-order optimization
technique.

%% file: method.tex

In this section, we describe three new intra-processing algorithms for debiasing 
neural networks. First we give more notation and
formally define the different types of debiasing algorithms.

Given a neural network $f_\theta$, where $\theta$ represents the weights, 
we sometimes drop the subscript $\theta$ when it is clear from context.
We denote the last layer of $f$ by $f^{(\ell)}$, and we assume that
$f=f^{(\ell)}\circ f'$, where $f'$ is all but the last layer of the neural network.
One of our three algorithms, layer-wise optimization, 
assumes that $f$ is feed-forward, that is, $f=f^{(\ell)}\circ\cdots\circ f^{(1)}$
for functions $f^{(1)},~f^{(2)},\dots,f^{(\ell)}.$
The performance of the model is given by a performance measure 
$\rho$.
For a set of data points $\mathcal{D}$, given the list of true labels $\mathcal{Y}$ 
and the list of predicted labels 
$\hat{\mathcal{Y}}=[f(\pmb{x}_i)\mid (\pmb{x}_i,Y_i)\in \mathcal{D}]$, 
the performance is $\rho(\mathcal{Y},\hat{\mathcal{Y}})\in [0,1].$
Common performance measures include accuracy, precision, recall,
or AUC ROC (area under the ROC curve). In our experiments we use balanced accuracy as our performance measure. The formal definition of balanced accuracy is

\begin{equation*}
    \rho(\Y, \hat \Y) =  \frac{1}{2} \left(\frac{\mathit{TPR}}{\mathit{TPR} + \mathit{FNR}} + \frac{\mathit{TNR}}{\mathit{TNR} + \mathit{FPR}} \right).
\end{equation*}
We also define a bias measure $\mu$, given as 
$\mu(\mathcal{D}, \hat {\mathcal{Y}}, A)\in [0,1]$,
such as one defined in Section~\ref{sec:prelim}.

The goal of any debiasing algorithm is to increase the performance $\rho$, 
while constraining the bias $\mu$.
Many prior works have observed that fairness comes at the price of accuracy
for many datasets, even when using large models such as deep 
networks~\citep{bellamy2018ai, verma2018fairness, chouldechova2017fair}.
Therefore, a common technique is to maximize the performance subject to a constraint
on the bias, e.g., $\mu<0.05$.
Concretely, we define an objective function as follows.
\begin{equation}
 \phi_{\mu, \rho, \epsilon}(\mathcal{D}, \hat{\Y}, A)  = \begin{cases} \rho \text{ if }\mu<\epsilon,\\
 0\text{ otherwise.}
\end{cases}
\label{eq:objective}
\end{equation}

An in-processing debiasing algorithm takes as input the training and validation
datasets and outputs a model $f$ which seeks to maximize $\phi_{\mu,\rho,\epsilon}$.
An intra-processing algorithm takes in the validation dataset and a trained model $f$
with weights $\theta$ (typically $f$ was trained to optimize the performance $\rho$ on the training dataset), 
and outputs fine-tuned weights $\theta'$ such that $f_{\theta'}$ maximizes
the objective $\phi_{\mu,\rho,\epsilon}.$
A post-processing debiasing algorithm takes as input the validation dataset
as well as a set of predictions $\hat{\Y}$ on the validation dataset
(typically coming from a model $f$ which was optimized for $\rho$ on the training dataset),
and outputs a post-processing function $h:[0,1]\rightarrow \{0,1\}$ which
is applied to the final output of the model so that the final predictions optimize
$\phi_{\mu,\rho,\epsilon}.$
Note that intra-processing and post-processing debiasing algorithms are useful in
different settings. Post-processing algorithms are useful when there is no
access to the original model. Intra-processing algorithms are useful when there is
access to the original model, or when the prediction is over a continuous feature.
Now we present three new intra-processing techniques.

\paragraph{Random perturbation.}
Our first intra-processing algorithm is a simple iterative random procedure, 
\emph{random perturbation}. In every iteration, each weight in the neural network 
is multiplied by a Gaussian random variable with mean 1 and standard 
deviation 0.1. In case the model $f$ outputs probabilities, we
find the threshold $\tau$ such that 
$\hat{\Y_\tau}=[\ind\{\hat Y_i>\tau\} | \hat Y_i\in \hat{\Y}]$ 
maximizes $\phi_{\mu,\rho,\epsilon}(\Y,\hat{\Y_\tau}, A).$
We run $T$ iterations and output the perturbed weights that maximize 
$\phi_{\mu,\rho,\epsilon}$ on the validation set.
See Algorithm~\ref{alg:random}.
We show in the next section that despite its simplicity,
this model performs well on many datasets and fairness measures,
and therefore we recommend this algorithm as a baseline
in future intra-processing debiasing applications.
A natural follow-up question is whether we can do even better by using
an optimization algorithm instead of random search.
This is the motivation for our next approach.

\begin{algorithm}
  \begin{algorithmic}[1]
    \STATE {\bfseries Input:} Trained model $f$ with weights $\theta$, 
    validation dataset $\mathcal{D}_{\text{valid}}$,
    objective $\phi_{\mu, \rho,\epsilon}$, parameter $T$
    \STATE Set $\theta^*=\theta,$ $\text{val}^*=-\infty$, and $\tau^* = 0$
    \FOR{$i=1$ to $T$}
         \STATE Sample $q_j\sim \mathcal{N}(1,0.1)$ for all $j\in \{1,2,...,|\theta|\}$
         \STATE $\theta'_j = \theta_j\cdot q_j$
        \STATE Select threshold $\tau\in [0,1]$ which maximizes the objective $\phi_{\mu,\rho,\epsilon}$ on the validation set
        \STATE Set $\text{val} =  \phi_{\mu,\rho,\epsilon}(\mathcal{D}_{\text{valid}}, \{\ind \{f_{\theta'}(\pmb{x}) > \tau\} \mid (\pmb{x}, Y) \in \mathcal{D}_{\text{valid}}\}, A)$
         \STATE If $\text{val}>\text{val}^*$, 
         set $\text{val}^* = \text{val}$, $\theta^*=\theta'$, and $\tau^*=\tau$.
    \ENDFOR
        \vspace{0.011in}
    \STATE {\bfseries Output:} $\theta^*$, $\tau^*$

\end{algorithmic}
 \caption{Random Perturbation}
 \label{alg:random}
\end{algorithm}

\paragraph{Layer-wise optimization.}
Our next method fine-tunes the model by debiasing individual layers using
zeroth order optimization.
Intuitively, an optimization procedure will be much more effective than
random perturbations. However, zeroth order optimization can be computationally expensive and does not scale well, 
so instead we only run the optimization on individual layers.
Given a model, assume the model can be decomposed into several functions
$f=f^{(\ell)}\circ\cdots\circ f^{(1)}$
For example, a feed-forward neural network with $\ell$ 
layers can be decomposed in this way.
We denote the trained weights of each component by 
$\theta_1,\dots,\theta_\ell$, respectively.
Now assume that we have access to a zeroth order optimizer $\mathcal{A}$,
which takes as input a model $f=f^{(\ell)}\circ\cdots\circ f^{(1)}$, 
weights $\theta=(\theta_1,\dots,\theta_\ell)$, dataset $\mathcal{D}_{\text{valid}}$,
and an index $i$.
The optimizer returns weights $\theta_i'$, optimized with respect to 
$\phi_{\mu,\rho,\epsilon}$.
In Algorithm~\ref{alg:deco}, we set the optimizer to be 
gradient-boosted regression trees (GBRT)~\cite{friedman2001greedy, mason2000boosting},
a leading technique for black box optimization which converts 
shallow regression trees into strong learners.
GBRT iteratively constructs a posterior predictive model
using the weights
to make predictions and uncertainty estimates for each potential set 
of weights $\theta'$.
To trade off exploration and exploitation, the next set of weights to try is
chosen using lower confidence bounds (LCB), a popular acquisition function
(e.g.,~\cite{joseph2016fairness}).
Formally, $\phi_\text{LCB}(\theta') = \hat{\theta'} - \beta \hat{\sigma},$ 
in which we assume our model's posterior predictive density follows a 
normal distribution
with mean $\hat{\theta'}$ and standard deviation $\hat{\sigma}.$
$\beta$ is a tradeoff parameter that can be tuned.
See Algorithm~\ref{alg:deco}.
Note that this algorithm can be easily generalized to optimize multiple
layers at once, but this comes at the price of runtime.
For example, running GBRT on the entire neural network
would be more powerful than the random permutation algorithm but is
prohibitively expensive.

 \begin{algorithm}
  \begin{algorithmic}[1]
    \STATE {\bfseries Input:} Trained model $f=f^{(\ell)} \circ \ldots \circ f^{(1)}$
    with weights $\theta_1,\dots,\theta_\ell$,
    objective $\phi_{\mu,\rho,\epsilon}$, black-box optimizer $\mathcal{A}$
    \STATE Set $\theta^*=\emptyset,$ $\text{val}^*=-\infty$, and $\tau^* = 0$
    \FOR{$i=1$ to $\ell$}
        \STATE Run optimizer $\mathcal{A}$ to optimize weights 
        $\theta_i$ to $\theta_i'$ with respect to $\phi_{\mu,\rho,\epsilon}$.
        \STATE Select threshold $\tau \in [0,1]$ which maximizes objective 
        $\phi_{\mu,\rho,\epsilon}$
        \STATE Set $\text{val} =  \phi_{\mu,\rho,\epsilon}(\mathcal{D}_{\text{valid}}, \{\ind \{f_{\theta'}(\pmb{x}) > \tau\} \mid (\pmb{x}, Y) \in \mathcal{D}_{\text{valid}}\}, A)$, where $\theta' = (\theta_1,...,\theta_i',...,\theta_\ell)$
        \STATE If $\text{val}>\text{val}^*$
         set $\text{val}^*=\text{val}$, $\theta^* = \theta'$, and $\tau^* = \tau$.
    \ENDFOR
    \STATE {\bfseries Output:} $\theta^*$, $\tau^*$
\end{algorithmic}
 \caption{Layer-wise optimization}
 \label{alg:deco}
\end{algorithm}

\paragraph{Adversarial fine-tuning.}
The previous two methods rely on zeroth order optimization techniques 
because most group fairness measures such as statistical parity difference
and equalized odds are non-differentiable. 
Our last technique casts the problem of debiasing as first-order optimization
by using adversarial learning.
The idea behind the adversarial method
is that we train a critic model to predict the amount of bias in a minibatch. We sample the datapoints in a minibatch randomly and with replacement. This statistical bootstrapping approach to creating a minibatch means that if the critic can predict the bias in a minibatch accurately, then it can predict the bias in the model with respect to the validation set reasonably well. Therefore, the critic effectively acts as
a differentiable
proxy for bias, which makes it possible to debias the original model using back-propagation.

The adversarial algorithm works by alternately iterating between training the
critic model $g$ using the predictions from $f$, and fine-tuning the predictive
model $f$ with respect to a custom function designed to be differentiable while still maximizing the non-differentiable objective function $\phi_{\mu,\rho,\epsilon}$ using the bias proxy $\hat{\mu}$ from $g$. The custom function we use to emulate the objective function while still being differentiable is given by $\max\{1, \lambda(|\hat{\mu}| - \epsilon + \mar) + 1\} \cdot Loss(y, \hat{y})$ where $\hat{y}$ is the predicted label, $y$ is the real label, $\lambda, \mar$ are hyperparameters, and $\text{Loss}$ is a generic loss function such as 
binary cross-entropy. This function multiplies the task specific loss by a coefficient that is 1 if the absolute bias is less than $\epsilon - \mar$, otherwise the coefficient is $\lambda(|\hat{\mu}| - \epsilon + \mar) + 1$. Intuitively, this custom function optimizes the task specific loss subject to the absolute bias less than $\epsilon - \mar$. The hyperparameter $\lambda$ describes how strict the bias constraint should be, and the hyperparameter $\mar$ describes the margin of error in the bias we want the algorithm to maintain. Note that the  $g$ concatenates the examples in the minibatch and returns a single number that estimates the bias of the minibatch as the final output. See Algorithm~\ref{alg:adversarial}. Note that BCELoss denotes the standard binary
cross-entropy loss.

\begin{algorithm}
  \begin{algorithmic}[1]
    \STATE {\bfseries Input:}
    Trained model $f=f^{(\ell)}\circ f'$ with weights $\theta$, validation dataset 
    $\mathcal{D}_{\text{valid}}$, parameters $\lambda,~\epsilon,~\mar,~n,~m,~m',~T$.
    \STATE Set $g$ as the critic model with weights $\psi$.
    \FOR{$i=1$ to $n$}
        \FOR{$j=1$ to $m$}
            \STATE Sample a minibatch $(\pmb{X}_j, \pmb {\Y}_j)$ with replacement from 
            $\mathcal{D}_{\text{valid}}$
            \STATE Evaluate the bias in the minibatch, $\bar \mu \leftarrow 
            \mu((\pmb{X}_j,\pmb {\Y}_j), f(\pmb{X}_j))$.
            \STATE Update the critic model $g$ by updating its stochastic gradient 
            $$\nabla_{\psi} (\bar \mu - (g \circ f')(\pmb{X}_j))^2$$
        \ENDFOR
        
        \FOR{$j=1$ to $m'$}
            \STATE Sample a minibatch $(\pmb{X}_j, \pmb {\Y}_j)$ with replacement from 
            $\mathcal{D}_{\text{valid}}$
            \STATE Update the original model by updating its stochastic gradient 
            $$\nabla_{\theta}\left[\max\{1, \lambda \cdot (|(g \circ f')(\pmb{X}_j)| - \epsilon + \mar)+1\} \cdot\text{BCELoss}(\pmb {\Y}_j, f(\pmb{X}_j)) \right]$$
        \ENDFOR
        \STATE Select threshold $\tau\in [0,1]$ that minimizes the objective $\phi_{\mu,\rho,\epsilon}$
    \ENDFOR
        \vspace{0.011in}
    \STATE {\bfseries Output:} Debiased model $f$, threshold $\tau$

\end{algorithmic}
 \caption{Adversarial Fine-Tuning}
 \label{alg:adversarial}
\end{algorithm}

\paragraph{Converting in-processing into intra-processing.}
Since both in-processing and intra-processing algorithms optimize the weights of the neural network while training, it may be possible to convert existing in-processing algorithms into intra-processing algorithms. 
However, the in-processing algorithm needs to be able to run on a generic neural architecture, and there cannot
be any specific weight-initialization step (since it is given a pretrained model as the starting point).
Furthermore, the hyperparameters of the in-processing algorithm may need to be modified to better fit the
fine-tuning paradigm. For example, the learning rate should be lowered, 
and the optimizer's influence on the earlier layers that are unlikely to contribute as much to the 
final result should be limited. 

As an instructive example, we modify a popular in-processing fairness algorithm~\citep{zhang2018mitigating} to convert it to the intra-processing setting. 
This algorithm relies on a similar adversarial paradigm to our adversarial fine-tuning algorithm.
The fundamental difference between the algorithm of~\citep{zhang2018mitigating} and our Algorithm~\ref{alg:adversarial} is that their algorithm uses the critic to predict the protected attribute and not to directly predict the bias for the minibatch. 
As a result, we modify our adversarial fine-tuning algorithm so the critic predicts the protected attribute, and we change the optimization procedure to the one provided by the original work. 
Finally we modify the hyperparameters so that they are better suited to the fine-tuning use case. For instance, we use a lower learning rate when fine-tuning the model.

%% file: experiments.tex
In this section, we experimentally evaluate the techniques laid out in
Section~\ref{sec:method} compared to baselines, on four datasets and
with multiple fairness measures.
To promote reproducibility, we release our code at
\url{https://github.com/abacusai/intraprocessing\_debiasing} and we use 
datasets from the AIF360 toolkit~\citep{bellamy2018ai} and a popular image dataset.
Each dataset contains one or more binary protected feature(s) and a binary label.
We briefly describe them below.

\paragraph{Tabular datasets.}
The COMPAS dataset is a commonly used dataset in fairness research,
consisting of over 10,000 defendants with 402 features~\citep{flores2016false}.
The goal is to predict the recidivism likelihood for an
individual~\cite{angwin2016machine}. 
We run separate experiments using \emph{race} and also 
\emph {gender} as protected attributes.
The Adult Census Income (ACI) dataset is
a binary classification dataset from the 1994 USA Census bureau database
in which the goal is to predict whether a person earns above \$50,000~\cite{Dua:2019}. 
There are over 40,000 data points with 15 features.
We use \emph{gender} and \emph{race} as the protected attribute.
The Bank Marketing (BM) dataset is from the phone marketing campaign of a Portuguese bank. 
There are over 48,000 datapoints consisting of 17 categorical and quantitative
features.
The goal is to predict whether a customer will subscribe to a product~\cite{moro2014data}.
The protected feature is whether or not the customer is older than 25.

\paragraph{The CelebA dataset.}
The CelebA dataset~\citep{celeba} is a popular image dataset used in computer 
science research. This dataset consists of over 200,000 images of celebrity head-shots, 
along with binary attributes such as ``smiling'', ``young'', and ``gender''. 
As pointed out in other papers~\citep{denton2019detecting}, the binary categorization
of attributes such as gender, hair color, and age, does not reflect true human diversity,
and is problematic~\citep{crawford2019excavating}.
Furthermore, some binary attributes present in CelebA such as ``attractive'' and 
``chubby'' involve pejorative judgements which may cause harm~\citep{denton2019detecting}.

In our experiments on CelebA, we choose to focus on two models. One model predicts whether or not the
person is classified as young, and the other predicts whether the person is classified as smiling. As pointed out in prior work, smiling detection has a host of potential 
positive applications with limited negative applications~\citep{denton2019detecting}.

We set the protected attribute to Fitzpatrick skin tones in range [4--6], as done in prior
work investigating inequity in computer vision 
models~\citep{wilson2019predictive, raji2020saving}.
The Fitzpatrick skin type scale~\citep{fitzpatrick1988validity} consists of six types of
skin tones, and generally tones 4--6 are darker than 1--3.
To label all 200,000 attributes, we used a pretrained classifier~\citep{zhang2020neural}.
We removed all images which were not predicted as type 1--3 or type 4--6 with at least 70\%
probability, which left us with roughly 180000 images.
Finally, we manually verified the correct classification of 1000 random images.

\paragraph{The need for neural networks.}
First, we run a quick experiment to demonstrate the need for neural networks
on the tabular datasets above.
Deep learning has become a very popular approach in the field of 
machine learning~\cite{lecun2015deep},
however, for tabular datasets with fewer than 20 features, it is worth checking
whether logistic regression or random forest techniques perform as 
well as neural networks~\cite{murphy2012machine}.
We construct a neural network with 10 fully-connected layers, BatchNorm for
regularization, and a dropout rate of 0.2, and we compare this to logistic
regression and a random forest model on the ACI dataset.
We see that a neural network achieves accuracy and
area under the receiver operating characteristic curve (AUC ROC) scores
which are 2\% higher than the other models.
See Appendix~\ref{app:experiments} for the full results.
Therefore, we focus on debiasing the neural network implementation.

\paragraph{Bias sensitivity to initial model conditions.}
We run experiments to compute the
amount of variance in the bias scores of the initial models.
Neural networks have a large number of local minima~\cite{swirszcz2016local}. 
Hyperparameters such as the optimizer
and learning rate, and even the initial random seed, cause the model to converge to different
local minima~\cite{lecun2015deep}. 
Techniques such as the Adam optimizer and early stopping with patience have
been designed to allow neural networks to consistently reach local minima with high 
accuracies~\cite{kingma2014adam, goodfellow2016deep}.
However, there is no guarantee on the consistency of bias across the local minima. In particular, the local minima found by
neural networks may have large differences in the amount of bias, and therefore, there may
be very high variance on the amount of bias exhibited by neural networks just because of the
random seed.
Every local optima corresponds to a different set of weights. If the weights of the model at a specific local optimum rely heavily on the protected feature, removing the bias from such a model by updating the weights would be harder than removing the bias from a model whose weights do not rely on the protected feature as heavily.
We compute the mean and the standard deviation of three fairness bias measures, as well as accuracy, 
for a neural network trained with 10 
different initial random seeds, across three datasets.
See Table~\ref{tab:variance}.
We see that the standard deviation of the bias score 
is an order of magnitude higher than the standard deviation of the accuracy.
In Appendix~\ref{app:experiments}, 
we also plot the contribution of each individual weight to the bias score, for a neural network. We show that the contribution of the weights to the bias score are sensitive to the initial random seed which means we cannot know the weights that are most likely to contribute to the bias before training the model even if we have another identical model trained using a different seed.

\begin{table}
\begin{center}
\caption{
\label{tab:variance}
Bias and accuracy of a neural network.}
\begin{tabular}{lllll}
\toprule
{} &                 AOD &                 EOD &                 SPD &         accuracy \\
\midrule
ACI (gender)  &  \hspace{-1mm}-0.084 $\pm$ 0.012 &  \hspace{-1mm}-0.082 $\pm$ 0.017 &  \hspace{-1mm}-0.198 $\pm$ 0.011 &  0.855$\pm$0.002 \\
BM (age)   &   0.011 $\pm$ 0.027 &  \hspace{-1mm}-0.009 $\pm$ 0.051 &   0.047 $\pm$ 0.015 &  0.901$\pm$0.002 \\
COMPAS (race) &   0.138 $\pm$ 0.017 &   0.194 $\pm$ 0.027 &   0.168 $\pm$ 0.016 &  0.669$\pm$0.006 \\
\bottomrule
\end{tabular}
\end{center}
\end{table}

\begin{figure*}
\centering %
\includegraphics[width=0.98\textwidth]{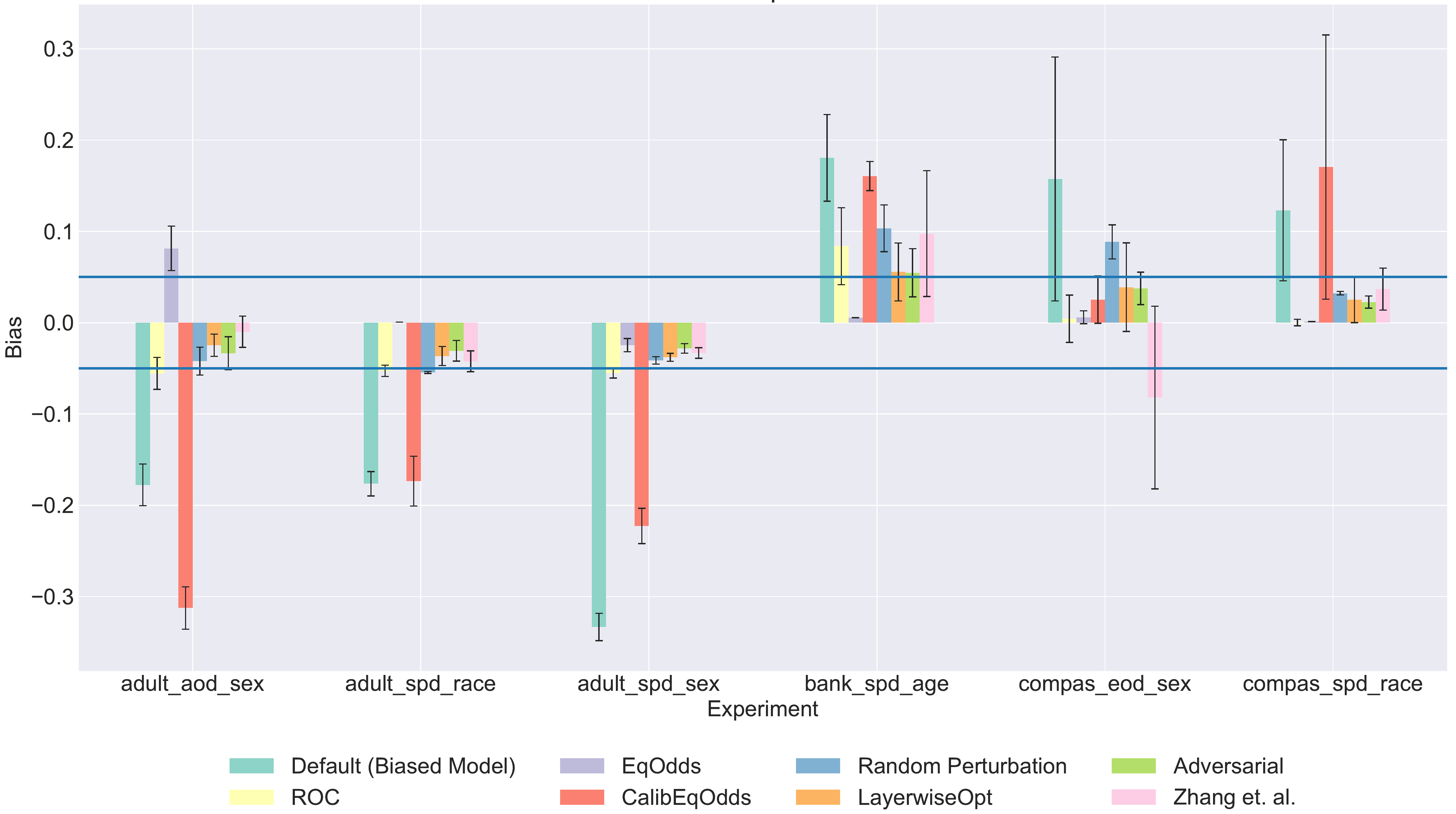}
\includegraphics[width=0.98\textwidth]{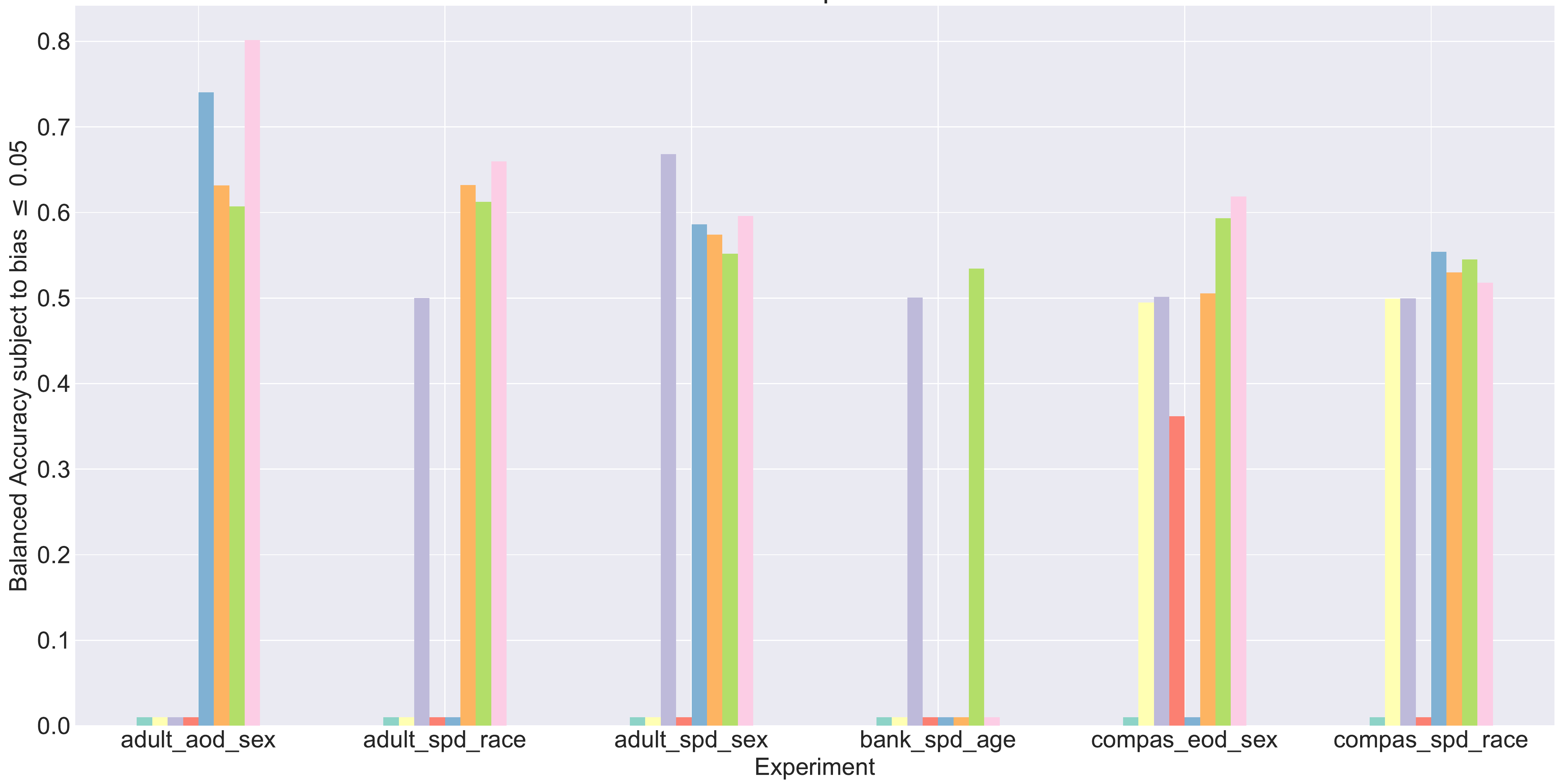}
\caption{
Results for tabular datasets over 5 trials. We plot the mean bias with std error bars (top),
and the median value of the objective function in Equation~\ref{eq:objective} (bottom).
Note that we report the median because Equation~\ref{eq:objective} has a large discontinuity.
}
\label{fig:posthoc_results}
\end{figure*}

\subsection{Intra-processing debiasing experiments}

Now we present our main experimental study by comparing four intra-processing and
three post-processing debiasing methods across four datasets and three fairness measures.
This includes one in-processing algorithm that we have adapted to the intra-processing
setting. First we briefly describe the baseline post-processing algorithms that we tested.

The \emph{reject option classification} post-processing 
algorithm~\citep{kamiran2012decision} defines a critical region of points in the 
protected group whose predicted probability is near $0.5$, and flips these labels.
This algorithm is designed to minimize statistical parity difference.
The \emph{equalized odds} post-processing algorithm~\citep{NIPS2016_6374}
defines a convex hull based on the bias rates of different groups,
and then flips the label of data points that fall inside the convex hull.
This algorithm is designed to minimize equal opportunity difference.
The \emph{calibrated equalized odds} post-processing algorithm~\citep{pleiss2017fairness}
defines a base rate of bias for each group, and then adds randomness based on the group into the classifier until the bias rates converge.
This algorithm is also designed to minimize equal opportunity difference.
For these algorithms, we use the implementations in the
AIF360 repository~\citep{bellamy2018ai}.

\begin{figure*}
\centering
\includegraphics[width=0.53\textwidth]{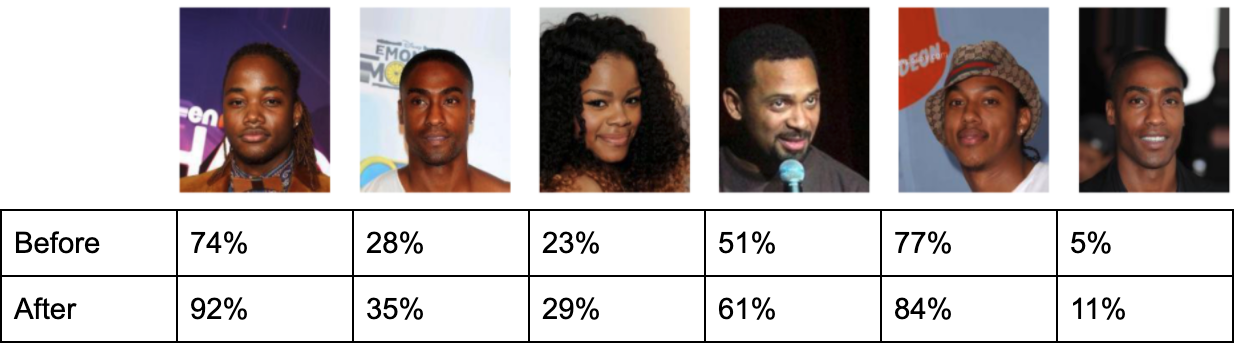}
\includegraphics[width=0.45\textwidth]{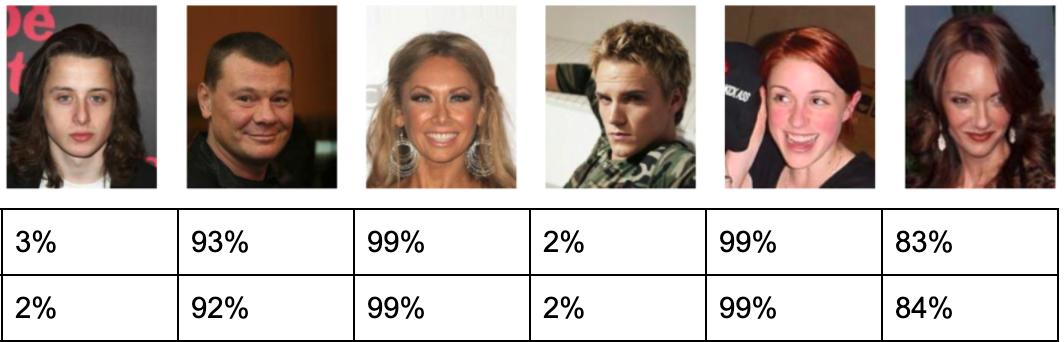}
\caption{Probability of smiling on the CelebA dataset, 
before and after debiasing w.r.t.\ race.
The adversarial fine-tuning method was used to debias.}
\label{fig:celeba}
\end{figure*}

Now we explain the experimental setup for the tabular datasets.
Our initial model consists of a feed-forward neural network with
10 fully-connected layers of size 32, with a BatchNorm layer between each fully-connected
layer, and a dropout fraction of 0.2. 
The model is trained with the
Adam optimizer 
and an early-stopping patience of 100 epochs.
The loss function is the binary cross-entropy loss.
We use the validation data as the input for the intra-processing methods,
with the objective function set to Equation~\ref{eq:objective}
with $\epsilon=0.05$.
We modified the hyperparameters so that each method took roughly 30 minutes.
We run each algorithm on 5 neural networks initialized with different
random seeds and aggregate the results. We publish the median objective scores for the tabular datasets as the mean would not accurately portray the expected results since some runs may return an objective score of 0.0. We also publish the mean and std error bars for the bias scores. We limit the plots to those for which the default biased algorithm did not achieve a positive objective score.
See Figure~\ref{fig:posthoc_results}. 

Finally, we run experiments on the CelebA dataset. We use a ResNet18 architecture~\citep{resnet}
pretrained on ImageNet from the PyTorch library~\citep{pytorch} as our initial model. 
Then we run 10 epochs of fine-tuning the architecture to predict the ``young'' attribute 
(or the ``smiling'' attribute) from the CelebA dataset using 40,000 random images.
To perform fine-tuning, we freeze all of the convolutional layers, only updating the final
fully-connected layers.
Then we run each of the debiasing algorithms.
We consider the default model, as well as the three post-processing algorithms and the three novel
intra-processing algorithms.
As in the previous section, we set the objective function to Equation~\ref{eq:objective}
with $\epsilon=0.05$. 
See Table~\ref{tab:celeba}. 
In order to give a qualitative comparison, 
we also publish a series of images along with the predicted probability that the celebrity is
smiling before and after applying the adversarial fine-tuning model.
See Figure~\ref{fig:celeba}.

\paragraph{Discussion.}
We see that the intra-processing methods significantly outperform the 
post-processing methods, sometimes even on the fairness metric for which the 
post-processing method was designed.
We note that there are two caveats. First, the three intra-processing methods had access
to the objective function in Equation~\ref{eq:objective}, while the post-processing
methods are only designed to minimize their respective fairness measures.
However, as seen in Figure~\ref{fig:posthoc_results}, sometimes the intra-processing
methods simultaneously achieve higher objective scores and lower bias compared to the
post-processing methods, making the intra-processing methods dominate them pareto-optimally.
Second, intra-processing methods are more powerful than post-processing methods, since
post-processing methods do not modify the weights of the original model.
Post-processing methods are more appropriate when the model weights are unavailable
or when computation time is constrained, and intra-processing methods are more appropriate
when higher performance is desired. 
We find that all the intra-processing algorithms tend to do well on the tabular datasets. However making sure that their bias scores remain below the threshold on the test is not as easy as there were not enough rows. Using regularization techniques helped to ensure that the scores remained consistent even over the test set.
We find that when the dataset and the model become more complex as is the case with the CelebA dataset and the ResNet model the more complex algorithms like adversarial fine-tuning tend to perform better than the random perturbation and layerwise optimization algorithm. This indicates that when dealing with more complex datasets and models, using complex intra-processing models like adversarial fine-tuning may be a better fit for the problem.

\begin{table}
\begin{center}
\caption{
\label{tab:celeba}
Results on the CelebA datasets for a pretrained ResNet with three initial random seeds. Results are the balanced accuracy scores after fine-tuning for the task of classifying whether the person in the image should be classified as young. The crossed out scores are those that did not have biases lower than 0.05.}
\begin{tabular}{lrrrrrrr}
\toprule
{} &  Default &     ROC &  EqOdds &  CalibEqOdds &    Random &  LayerwiseOpt &  Adversarial \\
\midrule
1 & \st{0.819} &  0.840 & \st{0.978} & \st{0.837} &  0.784 & 0.762 & 0.914 \\
2 & \st{0.823} &  \st{0.833} & \st{0.978} & \st{0.804} &  \st{0.777} &  0.889 & 0.917 \\
3 & \st{0.830} &  \st{0.852} & \st{0.977} & \st{0.837} &  0.801 & 0.750 & 0.905 \\
\bottomrule
\end{tabular}
\end{center}
\end{table}

%% file: conclusion.tex
In this work, we initiate the study of a new paradigm in debiasing research,
\emph{intra-processing}, which sits between in-processing and post-processing methods,
and is designed for fairly fine-tuning large models.
We define three new intra-processing algorithms:
random perturbation, adversarial fine-tuning, and layer-wise optimization,
and we repurpose a popular in-processing algorithm to work for intra-processing.
In our experimental study, first we show that the amount of bias is sensitive to the initial
conditions of the original neural network.
Then we give an extensive comparison of four intra-processing methods and three post-processing
methods across three tabular datasets, one image dataset, and three popular fairness measures.
We show that the intra-processing algorithms outperform the post-processing methods.

%% file: appendix_experiments.tex
\section{Additional Experiments and Details}\label{app:experiments}

In this section, we give additional details from the experiments
in Section~\ref{sec:experiments}, as well as additional experiments.

\paragraph{The need for neural networks.}
We start by comparing the performance of neural networks to logistic
regression and gradient-boosted regression trees (GBRT) on the datasets
we used, to demonstrate the need for neural networks.
This experiment is described at the start of Section~\ref{sec:experiments}.
For convenience, we restate the details here.
We construct a neural network with 10 fully-connected layers of size 32, 
BatchNorm for regularization, and a dropout rate of 0.2, and we compare this 
to logistic regression and GBRT on the ACI, BM, and COMPAS datasets.
See Table~\ref{tab:model_comp}. We see that the neural network
achieves better accuracy and ROC AUC on all datasets except COMPAS, 
which is within one
standard deviation of the optimal performance.

\begin{table}
\begin{center}
\caption{
\label{tab:model_comp}
Comparison between models. mean $\pm$ standard deviation}
\begin{tabular}{lllll}
\toprule
       &         & logistic regression &     neural network &      random forest \\
\midrule
ACI & accuracy &   0.852 $\pm$ 0.000 &  \textbf{0.855 $\pm$ 0.002} &  0.844 $\pm$ 0.002 \\
       & roc\_auc &   0.904 $\pm$ 0.000 &  \textbf{0.908 $\pm$ 0.001} &  0.889 $\pm$ 0.000 \\
BM & accuracy &  \hspace{-1mm}\textbf{ 0.901 $\pm$ 0.000} &  \textbf{0.901 $\pm$ 0.002} &  0.899 $\pm$ 0.001 \\
       & roc\_auc &   0.930 $\pm$ 0.000 &  \textbf{0.934 $\pm$ 0.001} &  0.932 $\pm$ 0.001 \\
COMPAS & accuracy &   \textbf{0.677 $\pm$ 0.000} &  0.641 $\pm$ 0.061 &  0.652 $\pm$ 0.006 \\
       & roc\_auc &  \hspace{-1mm}\textbf{ 0.725 $\pm$ 0.000} &  0.679 $\pm$ 0.088 &  0.695 $\pm$ 0.002 \\
\bottomrule
\end{tabular}
\end{center}
\end{table}

\paragraph{Bias sensitivity to initial model conditions.}
Next, we study the sensitivity of bias to initial model conditions.
Recall that in Table~\ref{tab:variance}, we computed the mean and standard
deviation of three fairness measures, as well as accuracy, for training a neural network
with respect to different initial random seeds. We see that standard deviation of the
bias is an order of magnitude higher than the standard deviation of the accuracy.
Now we run more experiments to show that the contribution of the weights to the bias score are sensitive to the initial random seed.

\begin{figure*}
\centering %
\includegraphics[width=0.51\textwidth]{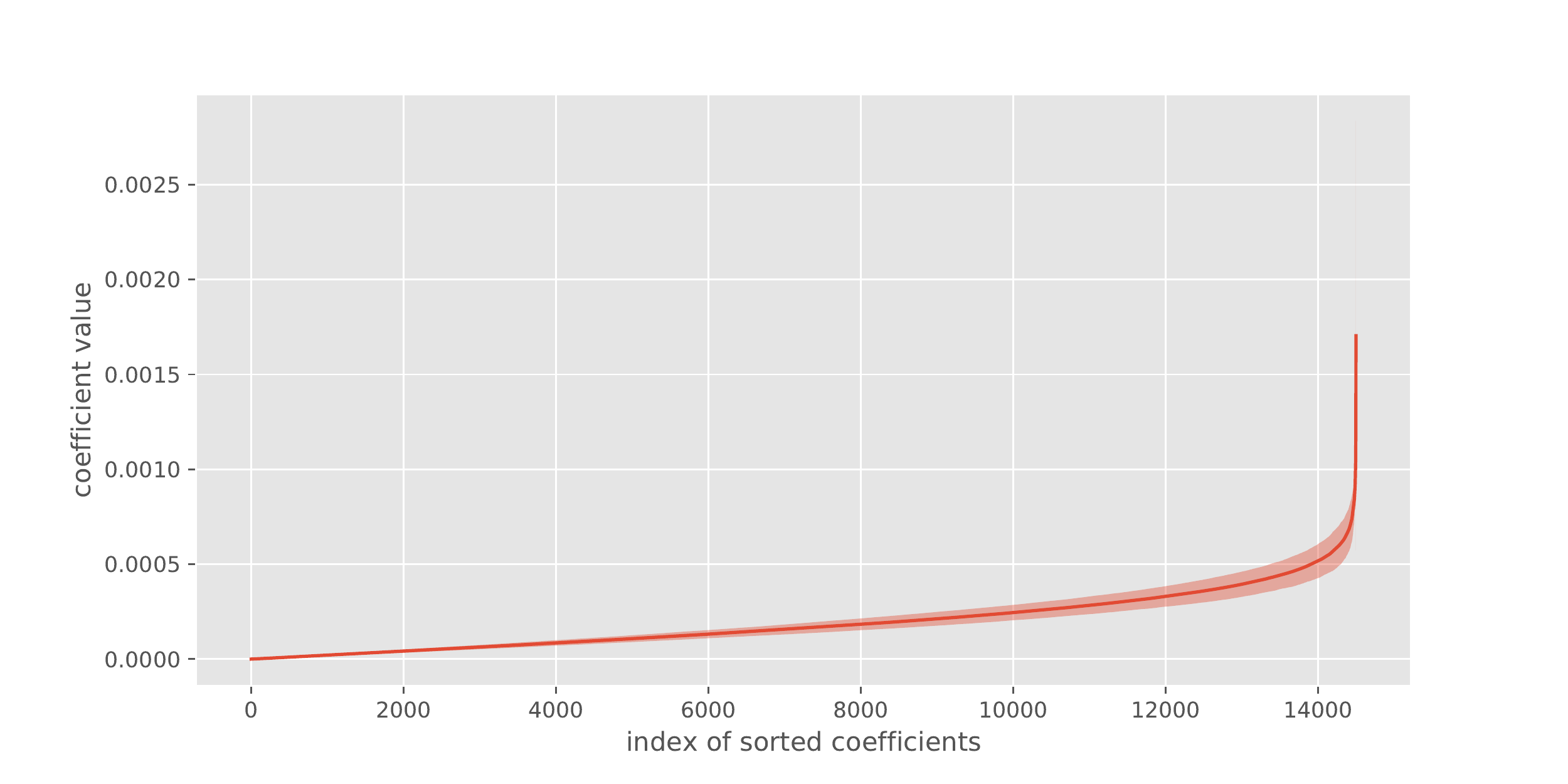}
\hspace{-.5cm}
\includegraphics[width=0.51\textwidth]{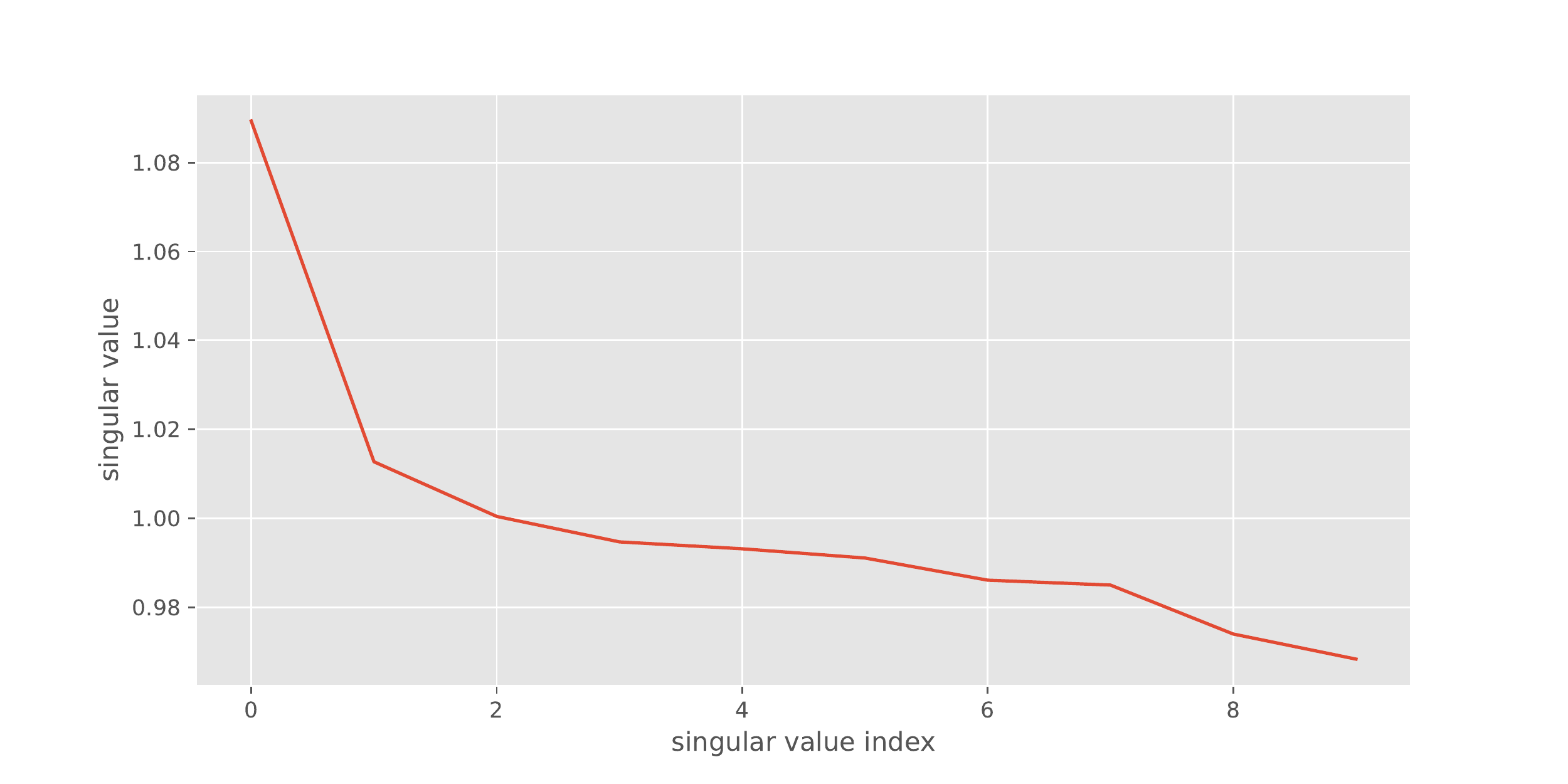}
\hspace{-.5cm}
\caption{
Results for coefficient analysis.
}
\label{fig:coefs_analysis}
\end{figure*}

For this experiment, we train 10 neural networks with the same architecture as described in Section~\ref{sec:experiments}. 
We want to identify which parameters of the network contribute most to the bias. To identify these parameters, we create 1000 random delta vectors with mean 1 and standard deviation 0.1 for each of the neural networks. We then take the Hadamard product of each random delta vector with the parameters of the corresponding network. We then evaluate the statistical parity difference (SPD) on the test set for the networks with the new perturbed parameters. To identify which parameters contribute most to the bias, we train a linear model for each of the 10 neural networks to predict the bias from the random delta vectors, and then we analyze the coefficients of the corresponding linear models. The linear models are successfully able to predict the bias based on the random delta vectors with an $R^2$ score of $0.861 \pm 0.090$. Figure~\ref{fig:coefs_analysis} (left) 
shows that only a small fraction of the parameters contribute to the majority of the bias. 

Now we want to identify how similar the coefficients of the linear models are across all 10 neural networks. To identify this, we stack the normalized coefficients for the linear models and decomposed the stacked matrix with singular value decomposition. The singular values of the matrix measure the degree of linear independence between the coefficients for the 10 linear models. As we see from Figure~\ref{fig:coefs_analysis} (right), the singular values are all close to 1. This indicates that the coefficients are all relatively different from one another. This means that the parameters of the 10 neural networks that correspond to the bias are different for each network indicating that each time we train a model, even if it has the same architecture, the parameters that contribute to bias are different. 

\begin{figure*}
\centering %
\includegraphics[width=0.98\textwidth]{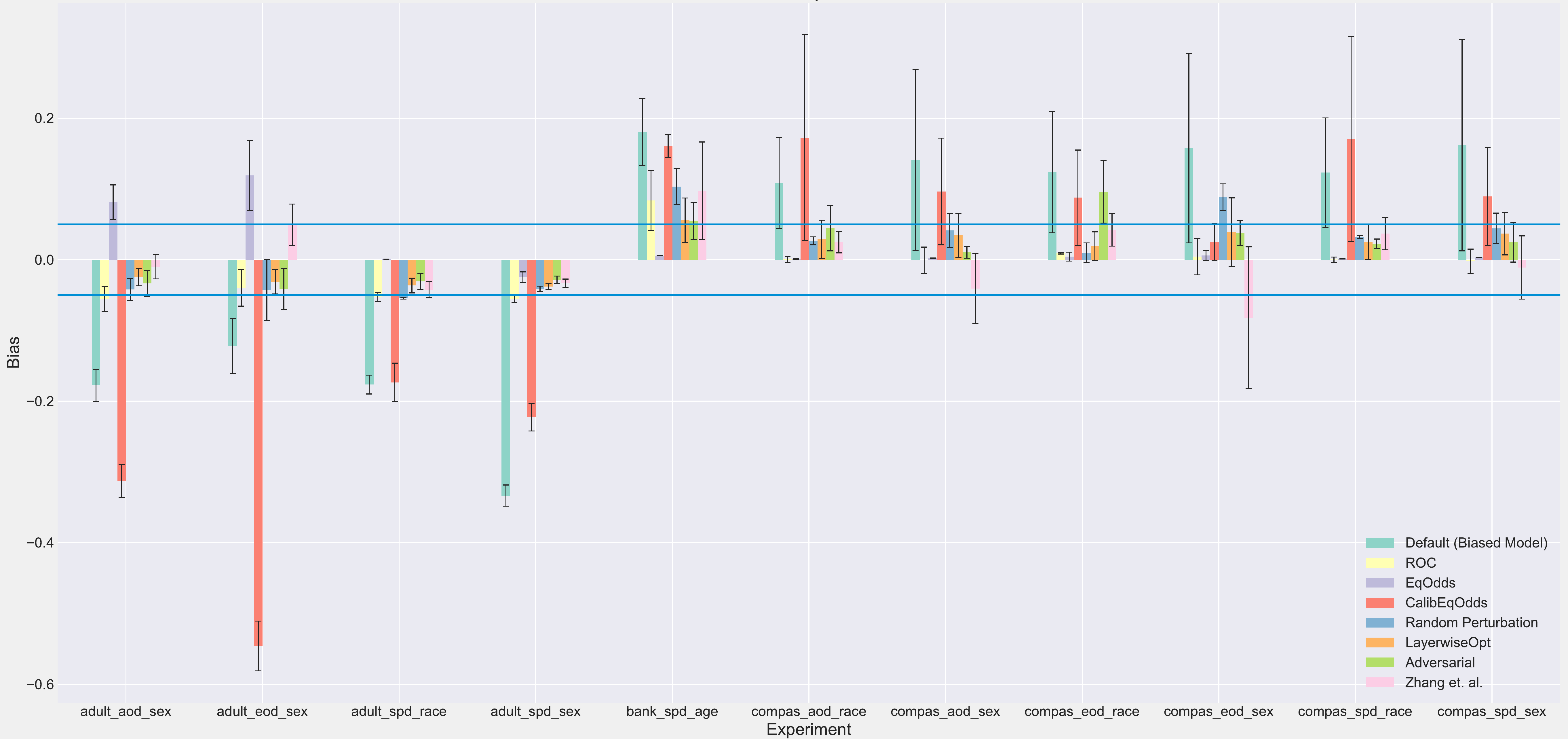}
\includegraphics[width=0.98\textwidth]{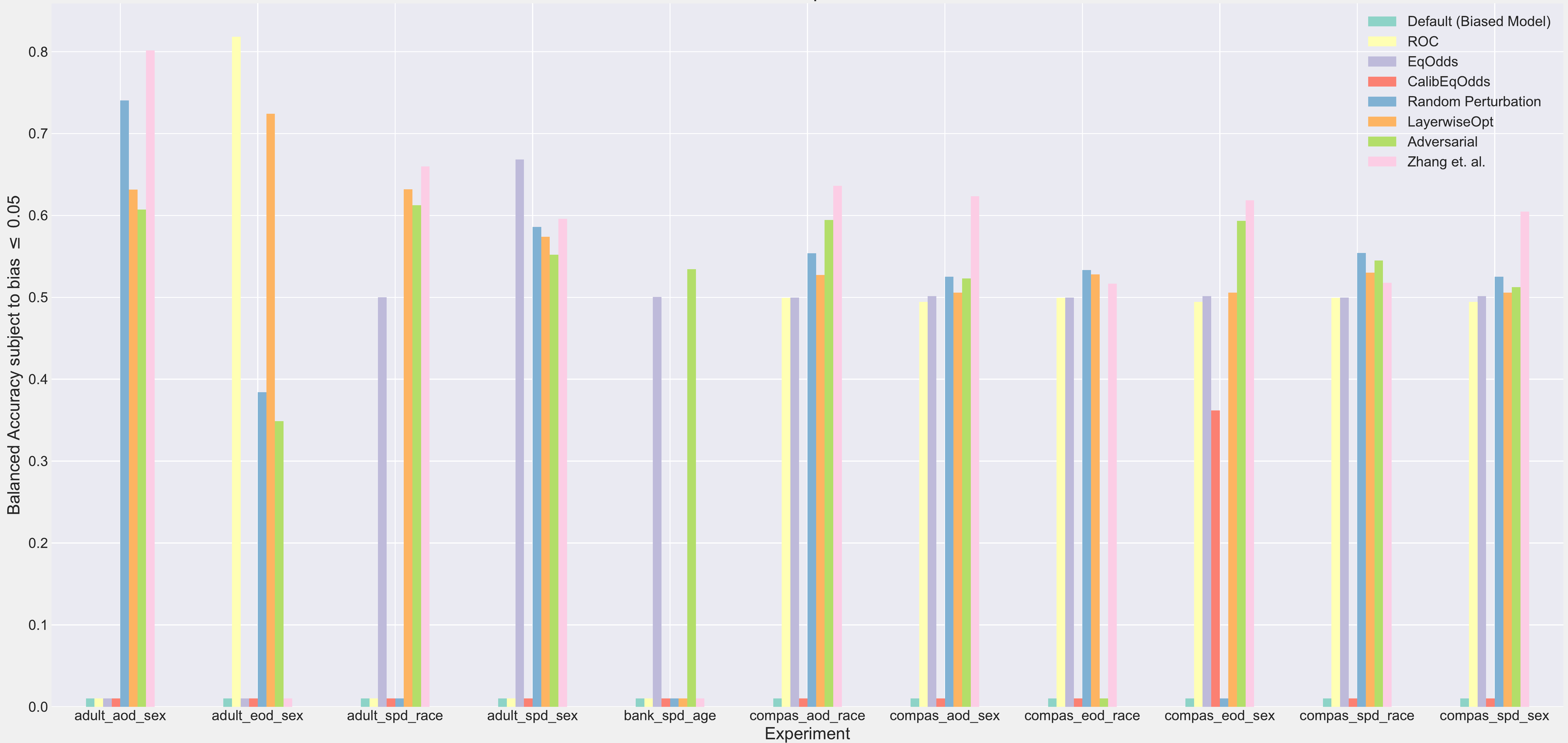}
\caption{
A continuation of the results from Figure~\ref{fig:posthoc_results}, with even more combinations
of dataset, bias measure, and protected attribute.
Over 5 runs with different seeds, we report
mean bias with std error bars (top) and the median of the objective function in Equation~\ref{eq:objective} (bottom).
Note that we report the median because Equation~\ref{eq:objective} has a large discontinuity.
}
\label{fig:additional_posthoc}
\end{figure*}

\paragraph{Tabular Intra-Processing Debiasing Experiments.}
Now we give more results for the intra-processing debiasing experiments from
Section~\ref{sec:experiments}.
The experimental setting is the same as in the tabular data experiments from
Section~\ref{sec:experiments} (Figure~\ref{fig:posthoc_results}).
We give even more combinations
of dataset, bias measures, and protected attributes.
See Figure~\ref{fig:additional_posthoc}.
Note that the results in this figure are a superset of the results from 
Figure~\ref{fig:posthoc_results}.
These additional results show that different debiasing measures work better in different situations.

